\documentclass[conference]{IEEEtran}

\IEEEoverridecommandlockouts
\usepackage[utf8]{inputenc}
\usepackage[T1]{fontenc}
\usepackage{graphics}
\usepackage{epsfig}
\usepackage{mathptmx}
\usepackage{amsmath}
\usepackage{amsfonts}
\usepackage{amssymb}
\usepackage{algorithm}
\usepackage{algpseudocode}
\usepackage{xcolor}
\usepackage{textcomp}
\usepackage{hyperref}
\usepackage{url}
\usepackage{xurl}
\usepackage[hyphenbreaks]{breakurl}
\usepackage{cleveref}
\usepackage{booktabs}
\usepackage{float}
\hypersetup{
    colorlinks=true,
    linkcolor=red,
    citecolor=blue,
    urlcolor=black,
    filecolor=black,
    linkbordercolor={0 0 0},
    pdfborder={0 0 0}
}
\def\BibTeX{{\rm B\kern-.05em{\sc i\kern-.025em b}\kern-.08em
    T\kern-.1667em\lower.7ex\hbox{E}\kern-.125emX}}

\begin{document}


\author{\normalsize Anita Srbinovska$^{1}$, Angela Srbinovska$^{1}$, Vivek Senthil$^{2}$,  Adrian Martin$^{3}$, John McCluskey$^{4}$, Jonathan Bateman$^{5, 6}$, Ernest Fokou\'e$^{7}$ \\
\small
$^{1}$Department of Computer Science, Rochester Institute of Technology, Rochester, NY, USA \\
$^{2}$School of Information, Rochester Institute of Technology, Rochester, NY, USA \\
$^{3}$Office of Business Intelligence, Rochester Police Department, Rochester, NY, USA \\
$^{4}$School of Criminal Justice, University at Albany, Albany, NY, USA \\
$^{5}$School of Individualized Study, Rochester Institute of Technology, Rochester, NY, USA \\
$^{6}$Department of Sociology and Anthropology, Rochester Institute of Technology, Rochester, NY, USA \\
$^{7}$School of Mathematics and Statistics, Rochester Institute of Technology, Rochester, NY, USA \\
\small Email: \{as2950, as2179, vs9589\}@rit.edu, Adrian.Martin@CityofRochester.gov, jmccluskey@albany.edu, \{jmb7342, epfeqa\}@rit.edu}


\title{Towards AI-Driven Policing: Interdisciplinary Knowledge Discovery from Police Body-Worn Camera Footage}

\maketitle
\thispagestyle{empty}
\pagestyle{empty}
\newcommand\tab[1][0.5cm]{\hspace*{#1}}


\begin{abstract}
    This paper proposes a novel interdisciplinary framework for analyzing police body-worn camera (BWC) footage from the Rochester Police Department (RPD) using advanced artificial intelligence (AI) and statistical machine learning (ML) techniques. Our goal is to detect, classify, and analyze patterns of interaction between police officers and civilians to identify key behavioral dynamics, such as respect, disrespect, escalation, and de-escalation. We apply multimodal data analysis by integrating image, audio, and natural language processing (NLP) techniques to extract meaningful insights from BWC footage. The framework incorporates speaker separation, transcription, and large language models (LLMs) to produce structured, interpretable summaries of police-civilian encounters. We also employ a custom evaluation pipeline to assess transcription quality and behavior detection accuracy in high-stakes, real-world policing scenarios. Our methodology, computational techniques, and findings outline a practical approach for law enforcement review, training, and accountability processes while advancing the frontiers of knowledge discovery from complex police BWC data.
\end{abstract}




\section{Introduction}

\begin{figure*}[h!]
    \centering
    \includegraphics[width=0.9\textwidth, keepaspectratio]{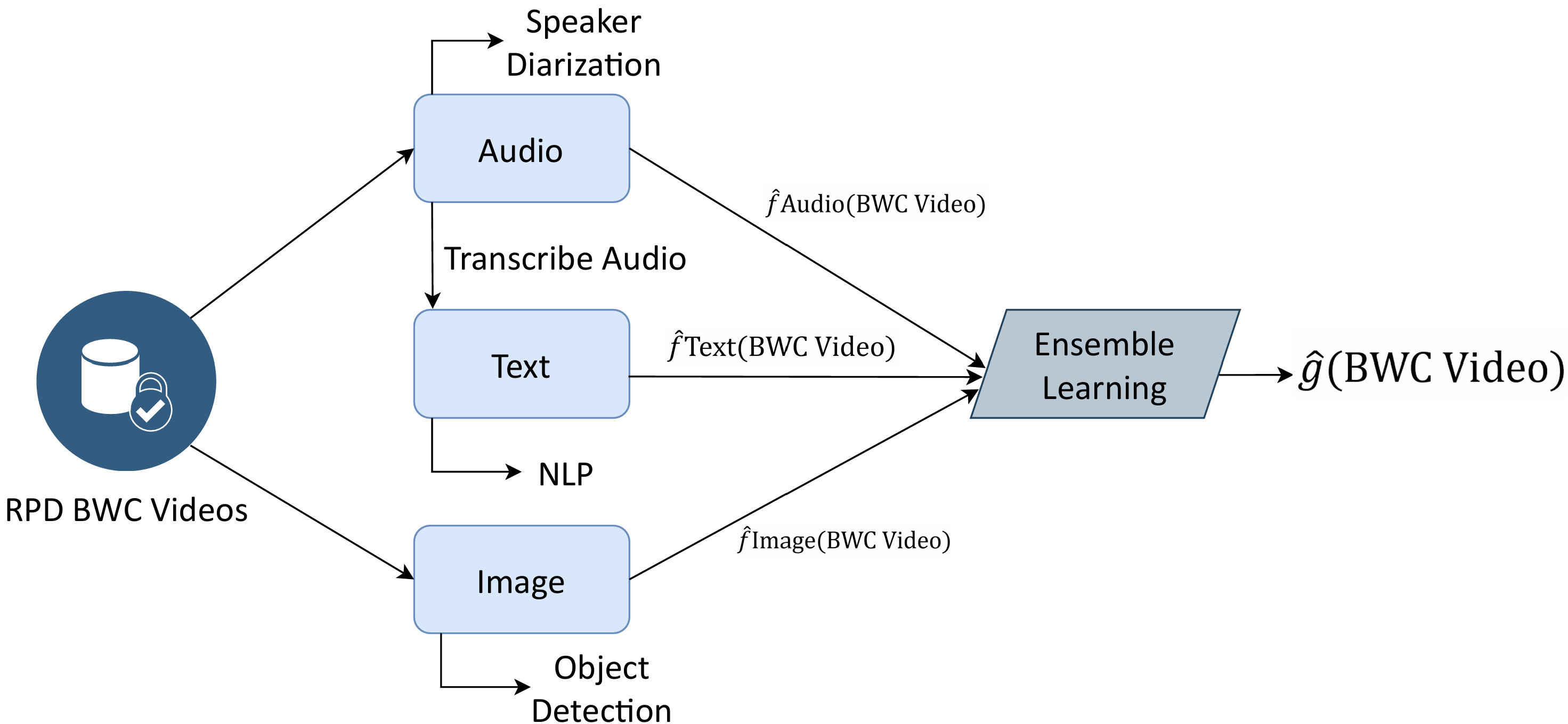}
    \caption{OpenBWC System Workflow: Multimodal Data Processing and Analysis Pipeline.}
    \label{fig:overview}
\end{figure*}
\noindent Police body-worn cameras (BWCs) are expanding video-gathering tools used for evidence collection that allow researchers, crime analysts, and academics to extract knowledge about police dynamics in real-time. The vast volume of data generated by BWC footage presents significant challenges across the criminal justice system, complicating efforts to analyze this information effectively and assess police-civilian interactions \cite{uchida2022}. To address these challenges, this paper introduces the \textit{OpenBWC}\footnote{\url{https://github.com/OpenBWC}} (Open Body-Worn Camera) framework, an open source, multimodal system designed to leverage artificial intelligence (AI) for analyzing patterns of behavior captured in BWC footage. The OpenBWC framework supports key goals in criminology by systematically analyzing respect, de-escalation, and escalation patterns in real-world police-civilian interactions, providing empirical evidence to inform training, oversight, and accountability. \\
\tab Using open source AI technologies in a responsible manner that minimizes bias, this framework aims to support informed decision-making in policing practices. Our approach integrates advanced speaker separation, audio transcription, and natural language processing (NLP) techniques to examine police-civilian interactions, enabling systematic analysis and classification of behavioral patterns. By applying these methods, we aim first to classify police-civilian interactions, focusing on key elements such as de-escalation, professionalism, and conflict resolution. This classification will help us identify the key factors that influence these outcomes. The goal is to translate analytical findings into practical strategies to improve police performance, with potential training approaches informed by collaboration with the Rochester Police Department (RPD). Current efforts focus on developing reliable methods to classify police-civilian interactions based on key behavioral and contextual elements. In this study, we look at how multimodal AI can help us analyze and categorize behavior in BWC footage. Our approach aims to improve police training, interactions between officers and civilians, and everyday policing work. We also consider the impact of civilian behavior or error, with the goal of advancing public safety, equity, and the broader pursuit of justice.


\section{Related Work}

\noindent Researchers have used police BWC footage as data to learn about policing for several years already (Makin et al., 2018) \cite{makin_et_al_2018}. With advancements in AI and NLP, this research has evolved to support large-scale analysis of recorded interactions. Computational methods have been used to examine officer behavior, communication strategies, and officer-civilian dynamics, making valuable contributions to policing research. \\
\tab Camp and Voigt (2025) \cite{camp_voigt_2025} proposed a framework that applies NLP techniques to BWC footage to identify behavioral patterns, assess communication styles, and evaluate policing practices. Their approach emphasizes scaling up analysis to highlight conversational themes, officer respect, and situational factors contributing to escalation or de-escalation. While they describe how supervised and unsupervised NLP models could extract linguistic features such as politeness, apology, and reassurance, their work remains largely theoretical, with the framework's practical implementation and performance yet to be demonstrated. While their framework is promising for automating behavior analysis, the authors emphasize that the system's accuracy may depend on factors such as the clarity of the transcripts, the reliability of the model, and the complexity of the conversations. They also note that background noise and different speaking styles could present additional challenges. These issues highlight the importance of thoroughly testing the framework in real-world police work. Even with these challenges, the framework may help us better understand patterns of interaction that are difficult to observe manually. \\
\tab As a related initiative, Camp et al. (2024) \cite{camp_pnas_2024} used BWC footage to assess the effects of communication-based training programs designed to improve officer behavior during traffic stops. Their study employed NLP models to analyze features such as respectful language, tone shifts, and conversational structure. Results indicated improved officer behavior and increased professionalism following targeted interventions, emphasizing the value of AI language models for training evaluation. \\
\tab Researchers have developed innovative annotation and labeling strategies to improve the scalability and performance of BWC footage analysis. One such tool is the Computer Vision Annotation Tool-Body-Worn Video (CVAT-BWV), a web-based platform designed to annotate BWC footage \cite{cvat_bwv_2024}. CVAT-BWV simplifies labeling footage with tools for marking key behaviors, conversation points, and non-verbal gestures. It combines automatic and manual labeling to support a more precise analysis of police-civilian interactions and to help identify important patterns in officer behavior. \\
\tab Moreover, recent efforts to improve the analysis of BWC footage have led to the development of video-based systematic social observation (VBSSO) methods. McCluskey et al. (2023) \cite{mccluskey2023} introduced VBSSO as a structured approach for studying police-civilian interactions using video data from police archives. Their method uses sampling strategies that connect BWC files to the department’s record management system (RMS), making it easier to select relevant data. These techniques have influenced the development of our framework and improved its ability to analyze behavior patterns in police-civilian interactions. \\
\tab Building on these foundations, we combine open source speaker separation models like SepReformer~\cite{shin2024} and transcription systems like WhisperAI \cite{radford2022} to assess the reliability of different tools in handling the challenging, noise-filled environments commonly found in BWC footage. Additionally, transcriptions were summarized using the open source Large Language Model (LLM) Llama 3.3 model \cite{touvron2024}. By integrating these tools with the open source database PostgreSQL for structured data organization, our system ensures scalable and reliable data retrieval to support comprehensive pattern analysis. This open source design improves accessibility and reproducibility, allowing researchers, machine learning (ML) practitioners, and police departments to adopt and adapt the framework for their investigations.


\section{Methodology}

\begin{figure}[H]
    \centering
    \includegraphics[width=0.95\columnwidth]{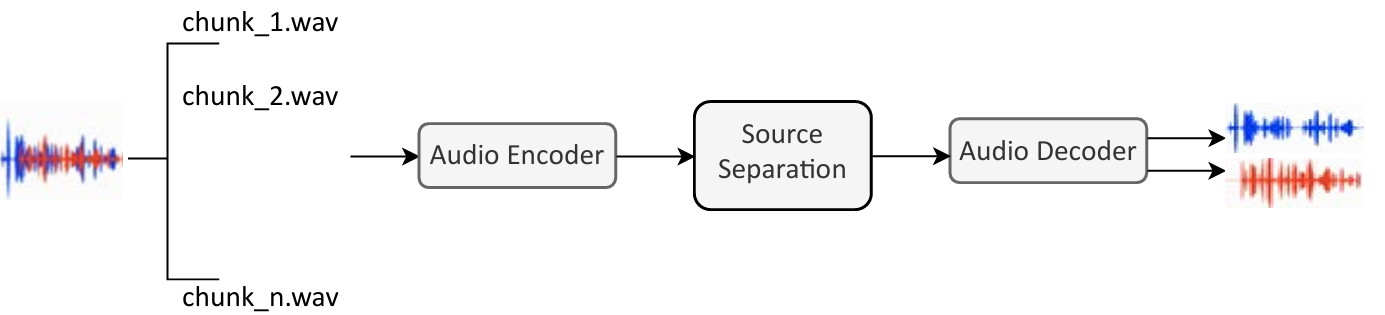}
    \caption{Multi-Speaker Audio Processing Flow.}
    \label{fig:speaker_pipeline}
\end{figure}

\noindent Our multimodal framework, as illustrated in Figure~\ref{fig:overview}, is designed to analyze audio, text, and image data in an integrated manner. To combine the strengths of each modality, the system applies an ensemble learning approach, described in \textit{Ensemble Learning for Multimodal Analysis }(Section~\ref{sec:ensemble_learning}), that integrates audio, text, and visual features into a unified analysis. This design uses quantitative and qualitative techniques to analyze behavioral patterns, conversational dynamics, and environmental context. The chosen methods emphasize data segmentation, advanced ML models, and structured data organization to ensure scalability and accuracy. \\
\tab At the core of the OpenBWC system is the \textit{knowledge extraction} algorithm (Algorithm~\ref{alg:knowledge_extraction}), which provides a step-by-step procedure for processing complex BWC footage. This algorithm extracts meaningful observations at each stage by leveraging advanced audio processing, transcription models, and NLP techniques. In parallel, knowledge extraction runs alongside every pipeline component to enrich the data with semantic labels and additional contextual information. \\ \\
\noindent \textbf{Data.} The dataset used in this research comprises 1,225 FOIL (Freedom of Information Law) videos obtained from publicly available sources through FOIL requests. The FOIL-requested videos have been redacted to blur faces and remove private or sensitive information before analysis. These recordings capture a broad spectrum of interaction contexts, including police-civilian conversations, emergency responses, and high-pressure situations. Each BWC recording, stored as an MP4 file, was systematically extracted from storage systems for further processing, including audio improvement, transcription, and structured data organization to enable systematic and thorough analysis. 
\begin{table}[h!]
    \renewcommand{\arraystretch}{1.1} 
    \small 
    \centering
    \caption{Summary of BWC FOIL Video Dataset}
    \label{tab:bwc_summary}
    \begin{tabular}{lc}
    \toprule
    \textbf{Measure} & \textbf{Value} \\
    \midrule
    Total Videos & 1,225 \\
    Shortest Video & 00:00:11 \\
    Longest Video & 11:59:30 \\
    Average Video Length & 1:31:59 \\
    Total Video Time & 1,877:50:01 \\
    \bottomrule
    \end{tabular}
\end{table}
\\ \tab Table~\ref{tab:bwc_summary} summarizes the BWC FOIL video dataset. Video durations range from 11 seconds to nearly 12 hours. On average, videos run approximately 1.5 hours, with over 1,877 hours of footage. For our test purposes in this phase, we broke down the videos into segments by officer BWC since FOILed videos of the same incident often come as a single video file that includes footage from multiple officers on the scene. This breakdown made our analysis simpler and more manageable.
\noindent \\ \\ \textit{Audio Processing Pipeline.} The audio pipeline was designed to handle long and complex recordings by applying chunking, speaker separation, and transcription models.
\begin{algorithm}[ht!]
    \small
    \caption{Knowledge Extraction from BWC Footage}
    \label{alg:knowledge_extraction}
    \begin{algorithmic}[1]
        \Procedure{AnalyzeFootage}{videoDataset}
            \For{each video in videoDataset}
                \State audio \(\gets\) ExtractAudio(video)
                \State videoChunks \(\gets\) SplitVideo(video, time)
                \For{each chunk in videoChunks}
                    \State speakers \(\gets\) SourceSeparation(chunk.audio)
                    \For{each speaker in speakers}
                        \State \parbox[t]{\linewidth}{transcript[speaker] $\leftarrow$ TranscribeAudio(\\speaker.audio)}
                    \EndFor
                    \State \parbox[t]{\linewidth}{mergedTranscript $\leftarrow$ MergeTranscripts(transcript)}
                    \State sceneObjects \(\gets\) SceneRecognitionAndObjectDetection(video)
                \EndFor
                \State \parbox[t]{\linewidth}{fullTranscript $\leftarrow$ SummarizeTranscripts(mergedTranscript)}
                \State VerifyAndCorrect(fullTranscript)
                \State insights \(\gets\) NLPAnalysis(fullTranscript)
                \State SaveInsights(insights)
            \EndFor
        \EndProcedure
    \end{algorithmic}
\end{algorithm}
\\ \tab The steps of this process are outlined in Algorithm~\ref{alg:knowledge_extraction}, which complements the description provided in this section. Following the algorithm alongside the narrative may help clarify the sequence of operations. In Line 1, the algorithm begins by iterating through the dataset of BWC videos. Each video undergoes audio extraction (Line 2), followed by segmentation into 15-second chunks (Line 3) during initial experiments. While this approach improved system responsiveness and reduced GPU load, it introduced challenges related to conversational continuity and maintaining context. Long audio files made the system work harder and slowed down the transcription process. To address this, we switched to 30-second chunks, which helped conversations sound more natural. This choice reflected a trade-off: shorter chunks often broke conversations into smaller parts, making it harder to maintain context and track speakers. In contrast, longer chunks better preserved conversational flow but required more computing power, making it harder to separate speakers during source separation. However, longer chunks may still be suitable in cases with fewer speakers or access to more powerful hardware. By working with these manageable audio units, the system could better capture conversational turns, improving speaker identification and transcription quality. \\
\tab Following segmentation (Lines 4-10), each audio chunk undergoes source separation using the SepReformer model—an asymmetric encoder-decoder architecture designed for noisy environments, as illustrated in Figure~\ref{fig:speaker_pipeline}. \\
\tab SepReformer is based on Blind Source Separation (BSS) principles, a technique that separates individual speaker signals from a mixture of overlapping audio sources without prior knowledge of the speakers. BSS models are particularly effective in noisy environments where traditional separation methods struggle. Inspired by Bando et al. (2024) \cite{bando2024neural}, our system uses SepReformer's advanced encoding-decoding structure to isolate individual speakers with improved accuracy. In this process, each audio chunk (e.g., chunk\_1.wav, chunk\_2.wav) is first encoded through an Audio Encoder, which extracts key audio features in the form of compact representations of the input sound that capture important patterns such as tone, pitch, and energy. These features, also called latent sequence features, help distinguish and separate speakers in a mixture. The Source Separation Module then processes the encoded features and isolates individual speakers from the mixed audio. Finally, the separated audio streams are decoded through the Audio Decoder to produce distinct waveforms corresponding to each speaker. The method outputs separate audio tracks for each participant, improving transcription accuracy and enabling more precise dialogue analysis in multi-party conversations.
\noindent \tab Once speaker separation was complete, transcribed segments were merged into a full transcript for each video (Line 12). This step utilizes WhisperAI (\texttt{base} model), optimized for real-world audio challenges such as noise, overlapping speech, and varied accents, common in BWC footage. \\
\tab After merging the transcripts in Line 10, the current implementation focuses on audio and text analysis for police-civilian interactions. However, future work will expand the algorithm to include scene recognition and object detection steps. These steps will analyze individual video frames to identify relevant objects (e.g., weapons, tasers, bystanders), estimate distances between officers and civilians, and extract contextual information about the environment and surroundings. \\
\tab In this context, accuracy goes beyond word-for-word correctness. It includes whether the transcription makes the conversation understandable and preserves the speaker’s emotional tone, timing, and intent, particularly in high-pressure moments. We observed that performance tended to be more reliable in routine interactions, such as traffic stops, where speech patterns are more predictable and audio is generally clearer. \\
\tab Based on our observations, challenging videos—such as those with chaos, shouting, or loud background noise—were harder to transcribe accurately. Common issues included repeated phrases, incorrect speaker labels, or missed tone shifts. These moments are important for evaluating system performance in high-stakes interactions, such as de-escalation. \\ \\
\noindent \textit{Text Analysis Pipeline.} The text analysis stage builds upon the transcribed audio by identifying conversational themes and structuring the data for retrieval. \\
\tab To ensure accurate speaker attribution, each transcript was aligned with the speaker-separated audio, which allowed us to determine whether the speaker was a police officer or a civilian. As part of our review, we randomly examined a subset of videos focused on routine traffic stops and standard police checks, where officers were typically the dominant speakers, often asking procedural questions, giving instructions, and assisting civilians. \\
\tab In Line 13 of Algorithm~\ref{alg:knowledge_extraction}, the full transcript is checked for alignment with the separated audio and metadata to verify speaker roles and maintain dialogue traceability. Once verified, Line 14 applies advanced NLP models, like Llama 3.3, to extract meaningful observations from each interaction, including de-escalation, tone, and respect indicators. \\ \\
\noindent \textit{Database and Retrieval System.} The processed results were organized in a PostgreSQL database to enable scalable data storage and efficient search. This structured database was designed to support rapid data retrieval and facilitate comprehensive pattern analysis. One of the key features of the database is its indexed summaries, which significantly enhance search speed and filtering efficiency. The system ensures that relevant data can be accessed quickly and accurately by indexing key details such as speaker identifiers and conversational themes. \\
\tab The database design was also built with scalability in mind. As the OpenBWC system continues to expand, this architecture can accommodate additional data sources, ensuring the system remains flexible and efficient as new records are integrated.

\subsection{Ensemble Learning for Multimodal Analysis}\label{sec:ensemble_learning}

\noindent To achieve comprehensive knowledge extraction from the BWC data, the OpenBWC system employs a mathematically driven ensemble learning framework designed to integrate diverse data features. By combining audio, text, and visual information, this method improves prediction accuracy. \\
\tab Let $\hat{f}_{\text{Audio}}$, $\hat{f}_{\text{Text}}$, and $\hat{f}_{\text{Image}}$ represent feature functions derived from audio, text, and image data, respectively. Each of these functions is defined as follows:
\[ \hat{f}_{\text{Audio}}(\text{BWC Video}) = h_{L} \circ h_{L-1} \circ \cdots \circ h_2 \circ h_1 (x) \]
Where:
\begin{itemize}
    \item $x$ is the original audio waveform from BWC footage. The length of the audio can impact processing; longer recordings (e.g., over 30 minutes) often led to speaker separation errors in our experiments, where the system began to confuse or switch speakers later in the interaction, especially during complex or overlapping conversations.
    \item Each transformation layer $h_i$ represents a feature extraction stage applied to the audio data, such as speaker diarization.
    \item The final transformation layer $h_L$ extracts high-level acoustic features, such as speaker-specific voice patterns.
\end{itemize}
\tab For the text analysis pipeline, the system extracts linguistic features from transcriptions using the following transformation function:
$$
\hat{f}_{\text{Text}}(\text{BWC Video}) = \text{Llama 3.3}\big(\text{WhisperAI}(x)\big)
$$
Where:
\begin{itemize}
\item $x$ is the original audio waveform extracted from the BWC footage.
\item WhisperAI transcribes the audio into text, while Llama 3.3 performs deep semantic analysis of the resulting transcription by identifying language markers such as politeness, de-escalation attempts, and reassurance cues.
\end{itemize}

\tab For visual analysis, the system defines:
\[ \hat{f}_{\text{Image}}(\text{BWC Video}) = v_{L} \circ v_{L-1} \circ \cdots \circ v_2 \circ v_1 (y) \]
Where:
\begin{itemize}
    \item $y$ is the sequence of video frames extracted from the BWC footage.
    \item Each layer $v_l$ represents a visual processing transformation, such as object detection and motion tracking.
\end{itemize}
\tab The final ensemble model combines these three feature representations to create a unified model. The integrated function is defined as:
\[ \hat{g}(\text{BWC Video}) = \alpha \hat{f}_{\text{Audio}} + \beta \hat{f}_{\text{Text}} + \gamma \hat{f}_{\text{Image}} \]
Where:
\begin{itemize}
    \item $\alpha$, $\beta$, and $\gamma$ are weighting coefficients that control the contribution of each data modality.
    \item The weighting values are calibrated during training to maximize the model’s accuracy.
\end{itemize}
\tab This ensemble model allows the OpenBWC system to exploit the complementary strengths of each data type—acoustic cues for speaker identification, textual information for dialogue analysis, and visual data for environmental context.


\section{Evaluations}

\subsection{Speaker Attribution and Role Labeling}

\noindent Speaker attribution was supported by the SepReformer model. However, in practice, we observed challenges, particularly in chaotic interactions or segments with multiple overlapping speakers, where the separation sometimes confused speakers or introduced artifacts. We compared the separated streams manually, noting that while officer speech was often reliably captured due to the microphone's proximity, civilian speech and overlapping dialogue were more challenging to separate accurately.

\subsection{Transcription Quality Assessment}

\noindent The selected tools—particularly WhisperAI—were tested on challenging BWC footage containing background noise, overlapping speech, and varied speaker dynamics. WhisperAI performed well on routine traffic stops and in less noisy environments. In high-stakes, chaotic scenarios, transcription accuracy dropped. Common issues included repeated phrases or missed key details, particularly concerning issues in oversight contexts. \\
\tab To systematically compare model performance, we developed a custom analysis pipeline in Python using the Natural Language Toolkit (NLTK) \cite{bird2009natural}. This pipeline analyzed transcripts produced by WhisperAI-\texttt{small} and WhisperAI-\texttt{base} through several steps. First, we tokenized the transcripts and calculated content coverage gaps, defined as the proportion of words present in one transcript but not in its counterpart. Next, we counted repeated lines, using a threshold of three or more repetitions to identify likely transcription artifacts. We also tallied non-standard characters that might indicate transcription errors or noise. Finally, we identified words not found in the English dictionary to measure potential linguistic distortions. \\
\tab We applied this analysis to 20 transcripts for a preliminary assessment. Results revealed that the \texttt{small} model consistently produced more transcription artifacts. Specifically, the content coverage gap for the \texttt{small} model averaged 34.3\%, indicating that a significant portion of the content from the \texttt{base} model was missing in the \texttt{small} transcript. Furthermore, the \texttt{small} model had an average of 549.0 repeated lines compared to 458.5 in the \texttt{base} model, suggesting a higher tendency for duplicated or looped phrases. Non-standard characters also appeared more frequently in the \texttt{small} model (4.10) than in the \texttt{base} model (3.35). \\
\tab It is important to note that the \texttt{base} model exhibited a higher average content coverage gap (39.6\%) than the \texttt{small} model. This is expected because the \texttt{base} model generally produces longer, more detailed transcripts, capturing a greater number of words and phrases. As a result, when comparing the \texttt{base} transcript to the \texttt{small} transcript, a larger portion of its content was not found in the smaller model’s output, highlighting that the \texttt{base} model was more comprehensive overall.
\begin{figure}[ht]
    \centering
    \includegraphics[width=\linewidth]{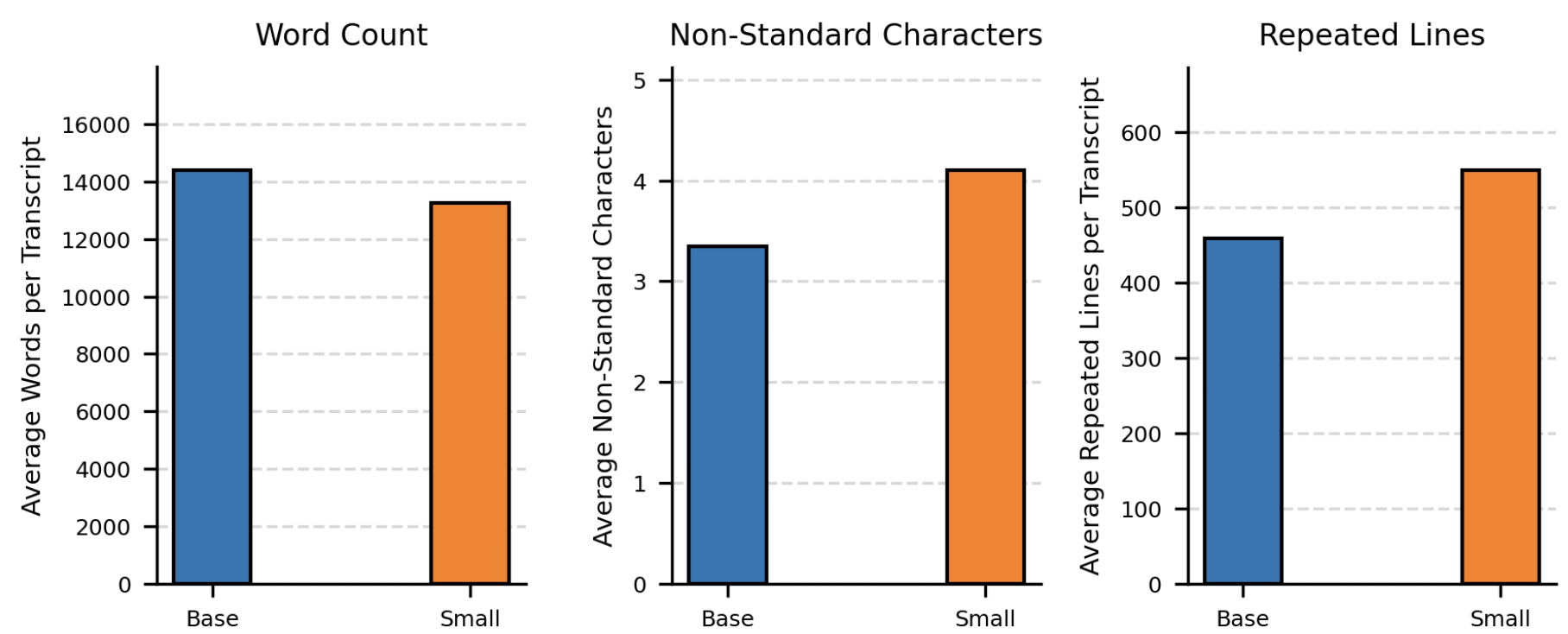}
    \caption{Comparison of transcription artifacts in the evaluation sample: (left) average word count per transcript; (center) average non-standard characters per transcript; (right) average repeated lines per transcript.}
    \label{fig:transcription_comparison}
\end{figure} \\
\tab Figure~\ref{fig:transcription_comparison} presents these results side by side, illustrating the relative performance differences between the two models. Based on these observations, we concluded that fully automated transcription pipelines using source separation and Whisper-\texttt{small} are insufficient for high-fidelity analysis of critical incidents. Consequently, we transitioned to using Whisper-\texttt{base} as our primary transcription source, supplemented by human verification and correction. This hybrid approach ensures more accurate transcripts that better reflect the original audio, supporting higher-quality downstream analyses. \\
\tab To support interpretability and thematic review, WhisperAI, SepReformer, and Llama 3.3 work alongside a structured data storage system. While these outputs aren't a replacement for human judgment, they help with initial analysis and review. This integration supports current analysis and lays the groundwork for future evaluations and replicable studies within the OpenBWC framework.


\section{Limitations}

\noindent Several challenges affected the system's performance and accuracy during development.
\begin{enumerate}
    \item \textit{GPU Memory Constraints.} Processing long audio recordings, especially those that span several hours, requires substantial GPU memory. The SepReformer model required extensive resources for such segments. To address this, the recordings were divided into 30-second chunks. RIT's Research Computing Services~\cite{rit_research} provided high-performance GPUs, enhancing processing speed and enabling efficient handling of large datasets.
    \item \textit{Noise Distortion.} BWC footage often contains significant background noise, including sirens, vehicle sounds, and overlapping speech. Although noise reduction methods made audio clearer, sometimes too much filtering removed important details, such as how a person sounded or subtle cues in their voice. While text transcriptions capture spoken words, vocal details such as tone and intonation convey important meaning, emotions, and intentions—details that are often lost with excessive audio filtering. Finding the right balance between reducing noise and keeping these vocal details is essential for accurately understanding noisy police-civilian interactions. Additionally, the officer’s voice is often captured with greater clarity than that of civilians, likely due to the proximity of the BWC's microphone and its orientation toward the officer’s speech. This asymmetry in audio capture can reduce the transcription quality for civilian speech in dynamic or noisy settings.
    \item \textit{Speaker Overlap.}
    Overlapping speech in BWC recordings reduced transcription accuracy. Although the SepReformer model effectively separated speaker audio when only two speakers were present, it struggled in chaotic scenarios characterized by rapid interruptions or when three or more speakers overlapped. This issue impacts transcription quality during precisely the incidents police managers prioritize for review—high-stakes, complex interactions. This finding aligns with previous research, such as Willits and Makin (2017) \cite{willits2018show}, who reported that most use-of-force cases recorded by BWCs lacked sufficient audio and video clarity for human coders to reliably interpret events. Improving audio separation techniques for multi-speaker situations represents an important area for future work, given their significance in understanding police-civilian interactions.
\end{enumerate}
\tab These technical limitations are particularly concerning in the context of public oversight and accountability, when clear and reliable audio is most important, like during high-stress or intense moments. Future work will aim to evaluate where, when, and why accuracy breaks down to inform both technical improvements and policy-level decisions about the appropriate use of BWC-derived evidence.


\section{Future Work}

\noindent Future work will use human-verified transcripts as the baseline for evaluating the system, rather than relying on automatic source separation, to improve transcription accuracy. The multimodal analysis will be refined to compare summaries of the same incident across multiple BWC videos, studying how consistent the interpretations are. We will include human-coded data to compare with automated outputs and to make the system more reliable and fair. \\
\tab Additionally, future work will expand the multimodal approach to include voice stress detection, sentiment analysis, and object detection to identify segments with emotional content and scene recognition to analyze the visual context, helping to capture a broader range of factors influencing police-civilian interactions. \\
\tab Collaboration with police officers will further support the interpretation of algorithm-clustered incidents by grounding findings in field expertise. \\
\tab These initiatives will help strengthen the alignment between algorithmic assessments and real-world policing challenges, supporting evidence-based improvements in police training and practice.


\section{Conclusion}

\noindent We introduced a framework demonstrating the potential of combining AI-driven audio, text, and image processing with advanced AI techniques for knowledge discovery from BWC footage. Analyzing repositories of BWC footage does not require access to proprietary software; instead, the tools can be open source, and the resulting algorithms, built to analyze publicly owned data, can also remain public. We identified key gaps in reliably assigning speaker identity during overlapping speech and maintaining transcription accuracy in high-noise, high-interruption scenarios, which introduce uncertainty in the transcribed content. Preliminary results support the value of the OpenBWC framework for interdisciplinary research, offering findings that can inform both criminological studies and practical applications in training, accountability, and policy development.

\section*{Acknowledgment}

\noindent This research is supported by a grant from the U.S. Department of Justice (DOJ) through its Office of Justice Programs (OJP) and Bureau of Justice Assistance (BJA), awarded to the City of Rochester.


\bibliography{references}

\end{document}